%% file: acl_latex.tex
\documentclass[11pt]{article}

\usepackage[final]{acl}

\usepackage{times}
\usepackage{latexsym}
\usepackage{booktabs}
\usepackage{multirow}
\usepackage{amsmath}
\usepackage{graphicx}
\usepackage{tikz}
\usepackage{xcolor}
\usepackage{natbib}
\usepackage{longtable}
\usepackage{array}
\usepackage{multirow}
\usetikzlibrary{positioning,arrows.meta,shadows.blur,calc}
\usepackage[table]{xcolor}
\usepackage{ragged2e}
\newcolumntype{L}[1]{>{\RaggedRight\arraybackslash}p{#1}}
\newcommand{\boundarysection}[2]{%
\midrule
\rowcolor{#1!12}
\multicolumn{5}{l}{\textbf{#2}}\\
\midrule
}
\usepackage[T1]{fontenc}

\usepackage[utf8]{inputenc}

\usepackage{microtype}

\usepackage{inconsolata}

\usepackage{graphicx}
\usepackage{xcolor}
\renewcommand{\thefootnote}{\fnsymbol{footnote}}

\title{Isolation as a First-Class Principle for LLM-Agent System Safety: Concepts, Taxonomy, Challenges and Future Directions}

\author {
    {\bf Huihao Jing}\textsuperscript{ \hspace{-0.2em}\includegraphics[height=1em]{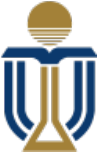}}\footnotemark[1],
    {\bf Wenbin Hu}\textsuperscript{ \hspace{-0.2em}\includegraphics[height=1em]{assets/HKUST.pdf}}\footnotemark[1],
    {\bf Shaojin Chen}\textsuperscript{ \hspace{-0.2em}\includegraphics[height=1em]{assets/HKUST.pdf}},
    {\bf Haochen Shi}\textsuperscript{ \hspace{-0.2em}\includegraphics[height=1em]{assets/HKUST.pdf}},
    {\bf Sirui Zhang}\textsuperscript{
      \hspace{-0.2em}
      \includegraphics[height=1em]{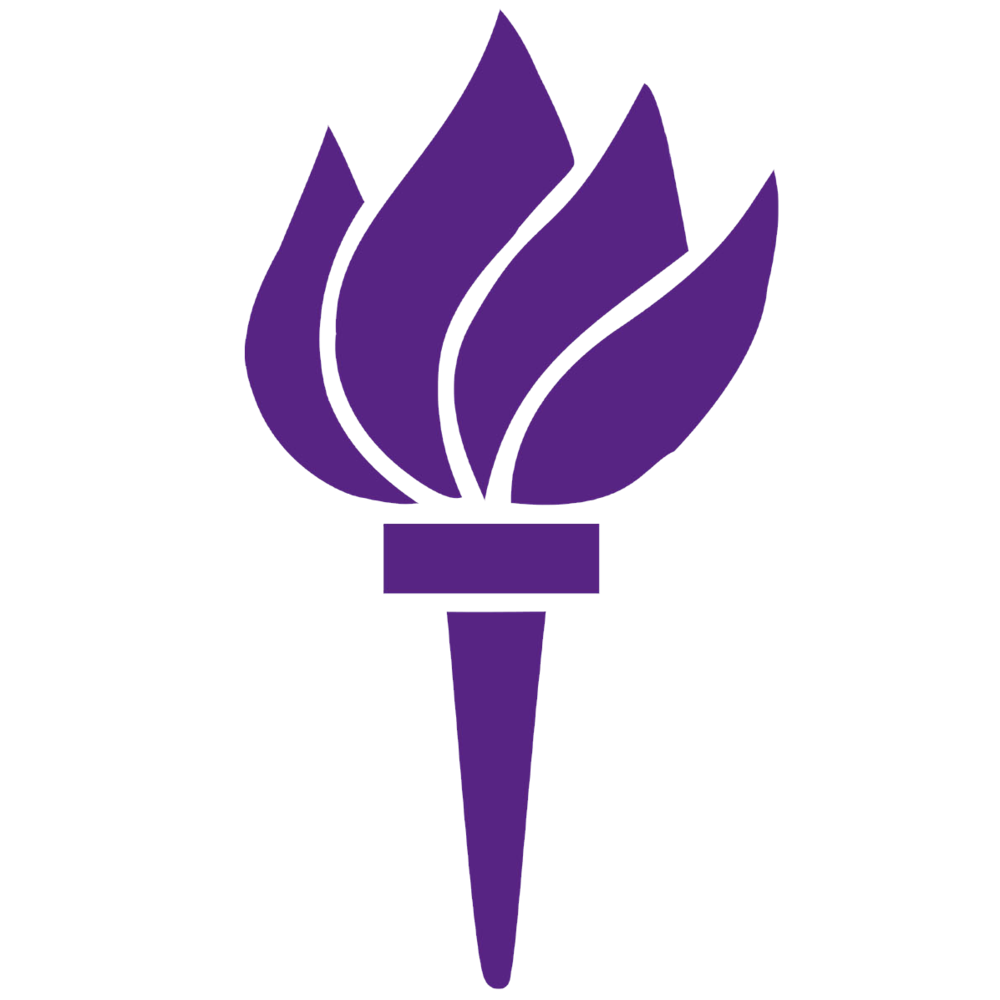}
      \hspace{0.1em}
      \includegraphics[height=1em]{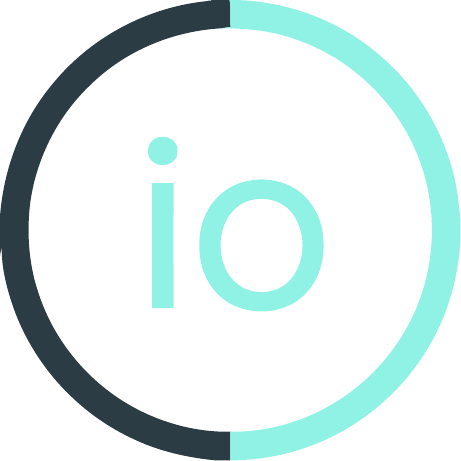}
    },\\
    {\bf Hanyu Yang}\textsuperscript{
      \hspace{-0.2em}
      \includegraphics[height=1em]{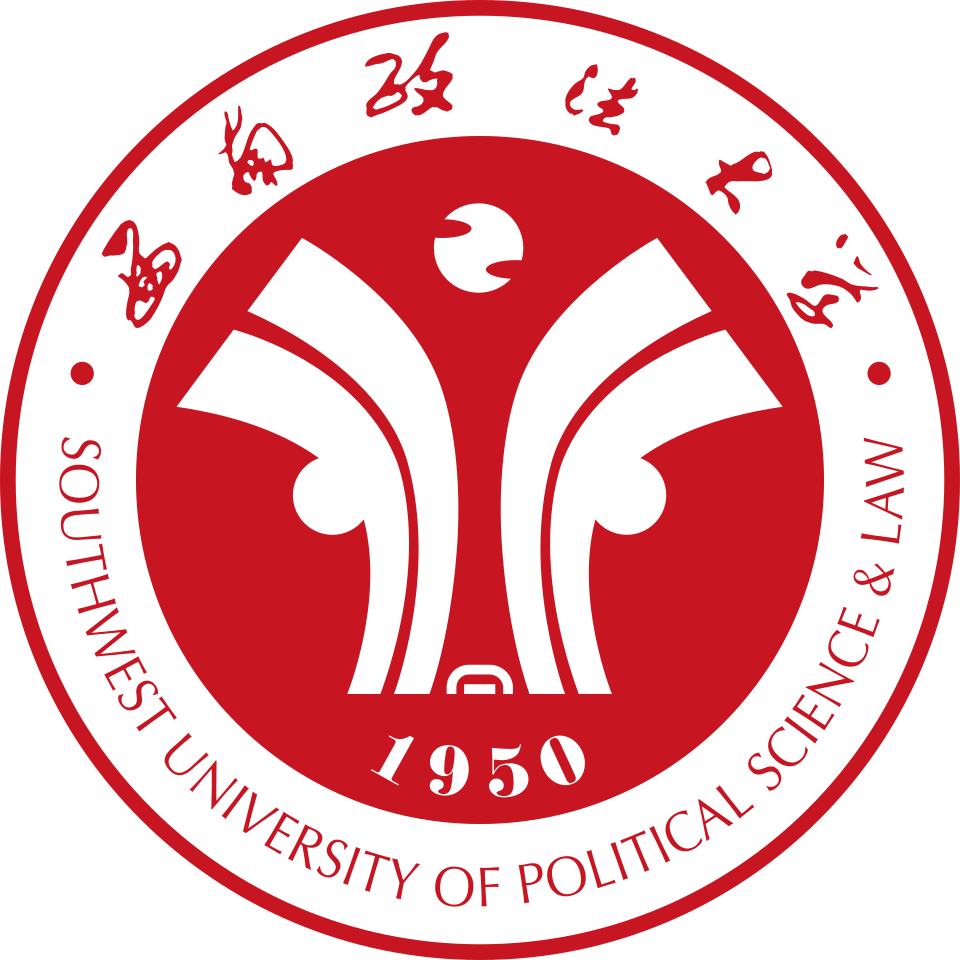}
      \hspace{0.1em}
      \includegraphics[height=1em]{assets/modeio.pdf}
    },
    {\bf Changxuan Fan}\textsuperscript{ \hspace{-0.2em}\includegraphics[height=1em]{assets/HKUST.pdf}},
    {\bf Zhongwei Xie}\textsuperscript{ \hspace{-0.2em}\includegraphics[height=1em]{assets/HKUST.pdf}},
    {\bf Hongyu Luo}\textsuperscript{ \hspace{-0.2em}\includegraphics[height=1em]{assets/HKUST.pdf}},\\
    {\bf Wun Yu Chan}\textsuperscript{ \hspace{-0.2em}\includegraphics[height=1em]{assets/HKUST.pdf}},
    {\bf Wei Fan}\textsuperscript{ \hspace{-0.2em}\includegraphics[height=1em]{assets/HKUST.pdf}},
    {\bf Haoran Li}\textsuperscript{
      \hspace{-0.2em}
      \includegraphics[height=1em]{assets/HKUST.pdf}
    }\footnotemark[2]\hspace{0.15em},
    {\bf Yangqiu Song}\textsuperscript{ \hspace{-0.2em}\includegraphics[height=1em]{assets/HKUST.pdf}}\\
    \textsuperscript{\includegraphics[height=1em]{assets/HKUST.pdf}}HKUST, 
    \textsuperscript{\includegraphics[height=1em]{assets/nyu.png}}NYU, 
    \textsuperscript{\includegraphics[height=1em]{assets/swupl.png}}SWUPL,
    \textsuperscript{\includegraphics[height=1em]{assets/modeio.pdf}}MODEIO.AI,\\
    \texttt{hjingaa@connect.ust.hk}\\
}

\begin{document}
\maketitle
\footnotetext[1]{Equal Contribution}
\footnotetext[2]{Corresponding author}
\renewcommand{\thefootnote}{\arabic{footnote}}
\begin{abstract}
The capability of LLM agents to function as the ``brain'' of a system fundamentally expands the scope of analysis beyond a standalone model. Consequently, safety is no longer only about input--output content alignment. It also concerns system behavior and real-world execution outcomes. However, the current literature is fragmented across attack types, applications, and benchmarks. This makes it hard to explain why failures such as prompt injection, tool misuse, and memory poisoning often share the same structural cause, and how they spread through an agent workflow. In this survey, we treat isolation as a first-class principle for LLM-agent system safety. By isolation, we refer to the separation of user inputs, tool access, execution channels, inter-agent communication, and environment-originated context. We organize the literature with a boundary-centric taxonomy of five boundaries: user-agent, agent-tool, agent-execution, agent-agent, and system-environment. This view helps identify where the loss of isolation first occurs, how compromise propagates across boundaries, and which defenses are most relevant at each interface. We also summarize cross-boundary failure paths, discuss open challenges, and outline a research agenda for isolation-by-construction in future agent systems.
\end{abstract}

\section{Introduction}

\begin{figure}
    \centering
    \includegraphics[width=1\linewidth]{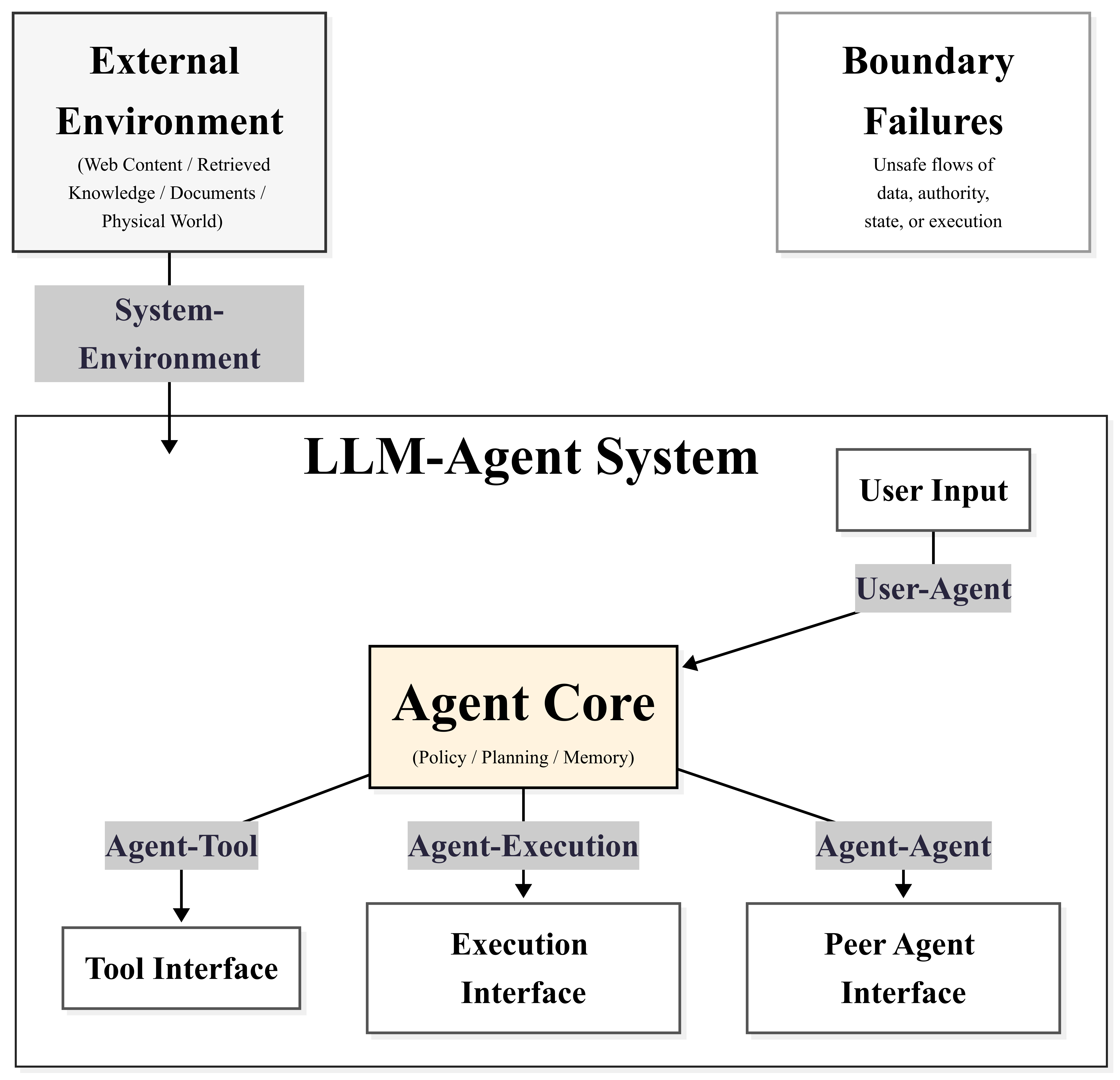}
    \vspace{-0.1in}
    \caption{A boundary-centric taxonomy of LLM-agent system safety organized around five isolation boundaries: user-agent, agent-tool, agent-execution, agent-agent, and system-environment.}
    \label{fig:agent-isolation}
    \vspace{-0.25in}
\end{figure}

LLM agents are moving quickly from research prototypes to real systems. Recent examples include Codex \cite{openai2025codex}, Claude Code \cite{anthropic2025claudecode}, and OpenClaw \cite{openclaw2026}. Related work on managed agents \cite{anthropic2026managed} and agent harnesses \cite{meng2026harness} shows the same shift. These systems do more than generate text. They read files, call tools, browse the web, edit state, coordinate with other agents, and take actions over long horizons. As a result, the safety problem is no longer only about unsafe outputs. It is also about how control, capability, memory, and context move through the system.

This change has already shaped both industry practice and recent research. A common direction is to decouple the brain from the hands. Managed agents \cite{anthropic2026managed}, AgentVisor \cite{ying2026agentvisor}, privilege-separation proposals \cite{jacob2025privilege}, CaMeL-style design defenses \cite{debenedetti2025defeating}, and Parallax \cite{fokou2026parallax} all push in this direction. Different papers use different terms, such as virtualization, privilege separation, typed interfaces, and design-level defense. But the core idea is simple: agent safety improves when system boundaries are explicit and enforced structurally, rather than left to prompt instructions alone.

Despite this progress, the literature remains fragmented. Existing surveys often organize the space by attack type, application domain, or agent capability \cite{meng2026harness,securityconsiderations2026,landscape2026}. That view is useful, but it leaves a gap. It does not clearly explain why failures that look different on the surface often share the same cause: loss of isolation at a boundary. It also makes cross-boundary propagation harder to see.

This survey addresses that gap. We treat isolation as a first-class principle for LLM-agent safety and organize the literature around five boundaries: user-agent, agent-tool, agent-execution, agent-agent, and system-environment. Figure~\ref{fig:agent-isolation} gives a compact view of this taxonomy. At the center is the agent core. Around it sit five interfaces where user inputs, tool access, execution channels, inter-agent communication, and environment-originated context may change status. The \textbf{user-agent boundary} concerns whether user content remains data or becomes control. The \textbf{agent-tool boundary} concerns how external capabilities are accessed. The \textbf{agent-execution boundary} concerns the transition from reasoning to action. The \textbf{agent-agent boundary} concerns communication and coordination across multiple agents. The \textbf{system-environment boundary} concerns how the agent system interacts with external content and state as a whole. These boundaries are distinct, but closely related. Together they provide a cleaner way to explain where compromise begins and how it later propagates. This taxonomy is boundary-centric, but not boundary-exclusive. Many papers naturally touch more than one boundary. We therefore classify papers by their \emph{primary safety boundary}, that is, the point where the loss of isolation first occurs. This rule keeps the taxonomy simple and makes cross-boundary analysis easier.

Our contributions are summarized as follows:
\begin{itemize}
\item We propose a boundary-centric taxonomy for LLM-agent safety based on five isolation boundaries, and map attacks/defenses to where isolation first breaks, showing how failures propagate across agent workflows.
\item We unify prior work through an isolation-centric view of failure propagation, and outline an isolation-by-construction agenda emphasizing trust separation, scoped capabilities, traceability, and recovery.
\end{itemize}

\input{texs/taxonomy}

\section{User-Agent Boundary}
\label{sec:user-agent}
\subsection{Threat Model and Boundary Definition}

The user-agent boundary asks whether an agent can keep user content from becoming privileged control. In a well-isolated system, user input should remain a request, a query, or task data. System and developer instructions should keep higher authority. Failure begins when user-controlled content starts to steer internal policy or behavior, as shown in early prompt-injection and role-confusion studies \cite{perez2022ignore,liu2023prompt,toyer2023tensor,pedro2023prompt,ren2024f2a, context_reasoner}. This is usually the first exposed interface in an agent system. Before an agent uses tools, acts, or coordinates with peers, it must decide what counts as instruction and what counts as data. If that decision is unstable, later safeguards start from a weak position.

\subsection{Direct Prompt Injection and Automated Jailbreaks}

Early work showed that natural-language inputs can override hidden prompts \cite{perez2022ignore,liu2023prompt}, leak system instructions \cite{toyer2023tensor}, and confuse role hierarchy \cite{pedro2023prompt}. The core issue is simple: a low-privilege source can behave like a high-privilege one. Recent work made this threat much more systematic. Attackers now use adversarial suffixes \cite{zhu2023autodan,liu2023autodan}, black-box search \cite{jia2024improved,sitawarin2024pal}, proxy-guided optimization \cite{jiang2024dartdeepadversarialautomated}, and automated red teaming \cite{pei2024selfprompt} to find failures at scale. Benchmarks such as HarmBench \cite{mazeika2024harmbench}, JailbreakBench \cite{chao2024jailbreakbench}, and JailbreakEval \cite{ran2024jailbreakeval} also pushed the field toward broader and more reproducible evaluation. Together, these papers show that the user-agent boundary is vulnerable under repeated adaptive pressure.

\subsection{From Single-Turn Attacks to Persistent Compromise}

More recent papers show that this boundary is not challenged only by single-turn prompts. Multi-turn jailbreaks exploit gradual steering \cite{cheng2024leveraging,russinovich2024great}, implicit clues \cite{ren2024derail,chang2024play}, and long-context overload \cite{upadhayay2024cognitive,zhou2024speak}. Many systems look safer in one-shot evaluation than they are in realistic sessions. The attack surface also expands in multimodal settings. User-provided images can carry visual or typographic instructions that bypass text-oriented safeguards \cite{liu2023query,gong2023figstep,wu2023jailbreaking}. This boundary is therefore not only about text. It spans multiple input channels. The strongest recent trend is persistence. In-context poisoning \cite{he2024data}, dynamic soft prompting \cite{wang2024unlocking}, and memory injection \cite{dong2025practical,xu2025mem} show that user influence can remain after the original interaction ends. Related work on poisoning alignment data \cite{shao2024making,zhang2024persistent} and harmful fine-tuning \cite{pathmanathan2024poisoning,huang2024antidote} shows that the same weakness can also enter through post-training pipelines. The problem is no longer only prompt-level override. It becomes long-term state corruption.

\subsection{Defenses, Evaluation, and Future Directions}

Defenses at this boundary fall into three broad groups. The first makes authority separation explicit through structured queries \cite{piet2023jatmo,chen2024struq}, signed prompts \cite{suo2024signed}, and DSL-style interfaces \cite{sharma2024spml}. The second hardens the model against jailbreaks through safety classifiers \cite{kim2023robust, GSPR}, semantic smoothing \cite{robey2023smoothllm}, repetition-based defenses \cite{ji2024defending,zhang2024parden}, and inference-time self-protection \cite{wang2024selfdefend,wang-etal-2024-self,lin2025uniguardian}. The third focuses on repair after degradation, including unlearning \cite{li2024backdoorllm,zhao2024survey}, editing \cite{lu2024eraserjailbreakingdefenselarge,zhao2024defendinglargelanguagemodels}, and refusal-boundary control \cite{xiong2024defensive,lu2025x,gou2024eyesclosedsafetyon}. Evaluation has also improved. Recent work studies detection with perplexity \cite{alon2024detecting}, refusal-loss signals \cite{hu2024gradient}, and geometry-aware methods \cite{candogan2025single,yung2025curvalid}. Benchmarks such as SG-Bench \cite{zhang2024sg}, Case-Bench \cite{sun2025casebench}, and Sorry-Bench \cite{xie2024sorry} broaden the test space beyond a few popular jailbreak prompts. Durability studies then ask a harder question: do safeguards still work under longer interaction and stronger adaptation \cite{qiEvaluatingDurabilitySafeguards2024a,tamirisaTamperResistantSafeguardsOpenWeight2025,chen2024characterizing, GrandGuard}?

Taken together, this literature suggests that user-agent safety must move beyond short prompts and static refusal scores toward long-session robustness, multimodal authority separation, and recovery from persistent compromise.

\section{Agent-Tool Boundary}
\label{sec:agent-tool}
\subsection{Threat Model and Boundary Definition}

The agent-tool boundary concerns how an agent accesses external capabilities. In a well-isolated system, tools should extend what the agent can do without taking over how it decides. Tool descriptions, tool outputs, and orchestration logic should remain constrained interfaces rather than hidden control channels. Failure begins when external capabilities are exposed without clear separation between observation, instruction, and execution, as shown by early indirect injection and tool-learning failures \cite{yi2023benchmarking,ye2024toolsword,fu2024imprompter}. 

This boundary matters because tools change the risk profile of an agent system. A standalone model can generate unsafe text. A tool-using agent can search the web, call APIs, read files, write code, or trigger downstream actions. Small control errors at this boundary can therefore turn into capability errors. The key question is not only whether the model reasons correctly, but whether the interface keeps capability use scoped, typed, and resistant to manipulation.

\subsection{Tool Outputs, Tool Misuse, and Tool Selection Failures}

The basic failure mode at this boundary is that tool-returned content is treated as trusted instruction. Early work on indirect prompt injection showed that hidden instructions in tool outputs can redirect downstream reasoning and action \cite{yi2023benchmarking}. Later work made this more concrete. ToolSword and Imprompter show that tool learning can fail at several stages, including interpretation, tool selection, argument generation, and feedback use \cite{ye2024toolsword,fu2024imprompter}. The result is not just poor task performance. It can also produce confidentiality, integrity, and safety failures. Recent papers push this line further by showing that manipulation can happen even before a tool is called correctly. Attacks on third-party APIs \cite{zhao2024attacks}, adversarial tool-calling \cite{wang2024allies}, and prompt injection against tool selection \cite{attacktoolselection2025} show that the agent can be steered into using the wrong capability, or using the right capability in the wrong way. Tool safety is therefore not one failure mode. The system can fail by choosing the wrong tool, passing unsafe arguments, trusting malicious output, or composing plausible calls into an unsafe workflow.

\subsection{Protocol, Metadata, and MCP-Style Risks}

A major recent shift is that tool-use risk is moving from content alone to protocol design. In MCP-style ecosystems, the model often sees tool descriptions, capability advertisements, and metadata before it uses the tool itself. MCIP and MPMA show that this layer is already security-critical \cite{jing2025mcip,wang2025mpma}. Metadata is not neutral from the model's perspective. It can shape preferences, change routing, and bias decisions before real tool output even appears. This is why recent benchmark work on MCP and tool ecosystems matters. MCP Security Bench \cite{msb2025} shows that attacks against model context protocol are not an edge case, but a natural extension of tool-mediated prompt injection. ToolSafe \cite{toolsafe2026} makes the same point from the defense side: tool invocation needs stronger safety mediation, not just better general alignment. The interface itself has become part of the attack surface. As tool use becomes standardized through protocols and orchestrators, the boundary no longer sits only at the moment of API invocation. It also sits at discovery, ranking, metadata interpretation, and workflow construction. Protocol semantics and capability exposure thus become first-class safety issues.

\subsection{Trajectory-Level Risk and Safe Tool Orchestration}

Recent work shows that tool attacks are often multi-step. A single call may look harmless, while the full trajectory becomes unsafe. AdapTools and TraceSafe make this clear in multi-step tool workflows \cite{adaptools2026,tracesafe2026}. The security target is therefore not only one prompt or one API call, but the full trace of calls, arguments, observations, and intermediate state. This is especially important for code interpreters and coding agents. CIBER and MOSAIC-Bench show that once tools support code execution or complex coding workflows, tool misuse can translate into direct downstream impact much more easily \cite{ciber2026,mosaic2026}. These systems are close to the execution boundary, but the first loss of isolation often still happens here, when a tool is selected, configured, or trusted too early. Defenses follow the same shift from call-level safety to interface-level safety. Some works harden the protocol layer through contextual integrity \cite{jing2025mcip}, metadata constraints \cite{wang2025mpma}, and benchmark-based auditing \cite{msb2025}. Others focus on safer invocation through least privilege, better routing, and orchestration-aware evaluation \cite{chen2025agentguard,toolsafe2026}. The broader lesson is simple: safer tool use depends less on the model inferring the right behavior from natural language, and more on interfaces that make capability scope, trust level, and action semantics explicit. This also points to future work on trace-level evaluation, privileged treatment of protocol metadata, and tighter links between tool control and execution control.

\section{Agent-Execution Boundary}
\label{sec:agent-execution}
\subsection{Threat Model and Boundary Definition}

The agent-execution boundary concerns the point at which internal decisions become real actions. In a well-isolated system, planning and action should remain separate enough that unsafe decisions can still be checked, delayed, or blocked before they create side effects. Failure begins when the system treats model output as ready-to-execute behavior without enough mediation.

This boundary turns control errors into operational impact. A model that produces unsafe text is one kind of problem. An agent that clicks the wrong button, runs unsafe code, submits the wrong form, or issues unsafe physical commands is another, as recent cyber-agent work makes clear \cite{zhang2024breaking,fang2024llma,fang2024llmb,guo2024redcode}.

\subsection{Code, Browser, and GUI Action Risks}

Recent work on cyber and software agents makes this transition clear. Agents can be redirected into malfunction amplification \cite{zhang2024breaking}, risky code generation \cite{fang2024llma}, or direct offensive behavior such as website compromise and vulnerability exploitation \cite{fang2024llmb,guo2024redcode}. The same pattern appears in browser and GUI agents. ST-WebAgentBench \cite{shlomov2024st} and SafeArena \cite{lee2025safearena} show that web agents can produce unsafe outcomes through clicks, submissions, navigation, and action grounding. Browser-agent work also shows that refusal-trained models remain vulnerable once they act through an interface rather than a chat window \cite{kumar2025refusal}, while GUI-agent studies show added risk from visual grounding, interface ambiguity, and hidden action consequences \cite{chen2025obvious}. Execution safety therefore cannot be reduced to ordinary alignment.

\subsection{Embodied Agents and Vision-Language-Action Systems}

The execution boundary becomes even sharper in embodied systems, where failures are often costlier and less reversible. BADROBOT \cite{zhang2024badrobot}, robot jailbreaking \cite{robey2024jailbreaking}, and contextual backdoor attacks \cite{liu2024compromising,jiao2025canwetrust} show that benign-looking inputs can induce unsafe physical behavior or persistent actuation failures. Benchmarks such as Embodied Agent Interface \cite{li2024embodiedagentinterface}, VIVA \cite{hu2024viva}, HASARD \cite{tomilin2025hasard}, HEAL \cite{chakraborty2025heal}, and Embodied Red Teaming \cite{karnik2025embodiedredteaming} make this risk more concrete. Work on VLA vulnerabilities further shows that execution can be corrupted through perception, including visual perturbations \cite{wu2024adversarial}, adversarial patches \cite{wang2025exploringadversarial}, and action-model weaknesses \cite{cheng2024manipulationfacingthreats,zhou2025exploringlimits}.

\subsection{Containment, Runtime Mediation, and Future Directions}

Because failures at this boundary create real side effects, defenses increasingly focus on containment rather than only prevention. Recent work proposes constrained execution \cite{wang2025advancingembodied}, zero-trust architectures \cite{lu2025poex}, active defense \cite{zhou2024haicosystem}, and policy-executable safeguards \cite{2603.17419}. Evaluation reflects the same shift. SafeArena \cite{lee2025safearena}, ST-WebAgentBench \cite{shlomov2024st}, and HWE-Bench \cite{hwebench2026} ask whether the agent remains safe while interacting with real interfaces and workflows. Overall, this boundary suggests that safe agent design must treat execution as a governed interface, not as the automatic continuation of model output.

\section{Agent-Agent Boundary}
\label{sec:agent-agent}
\subsection{Threat Model and Boundary Definition}

The agent-agent boundary concerns what happens when multiple agents communicate, delegate, debate, and share intermediate state. In a well-isolated system, one agent's message should remain a bounded contribution rather than an unverified control signal for others. Failure begins when inter-agent communication is treated as trusted reasoning by default. Multi-agent systems add channels that do not exist in single-agent settings. Messages can be forwarded, summarized, amplified, and stored in shared memory. As a result, one local failure may become a system-level failure, as shown by prompt-infection, malicious-agent, and propagation studies \cite{lee2024prompt,wang2024badagent,he2025red,wang2025g}.

\subsection{Prompt Infection and Communication Attacks}

The most direct failure mode at this boundary is communication-borne prompt injection. \emph{Prompt Infection} showed that one compromised agent can pass malicious instructions to others, turning ordinary coordination into an attack channel \cite{lee2024prompt}. A message is not just information. It can also be a carrier of control. Recent work shows that this problem becomes stronger in realistic collaborative settings. Debate-based attacks \cite{amayuelas2024multiagent}, manipulated-knowledge flooding \cite{ju2024flooding}, and explicit communication attacks \cite{he2025red} all show that harmful content can spread through discussion, critique, and consensus formation rather than through a single direct override. CORBA \cite{zhou2025corba} pushes this further by studying recursive blocking behavior, where one malicious pattern can propagate and suppress useful coordination across the network. The progression is clear: first one compromised message, then repeated propagation, then communication-level cascade.

\subsection{Malicious Agents, Topology, and Cascade Failures}

Another major line of work asks why some multi-agent systems fail locally while others fail systemically. One answer is topology. Network structure, routing rules, and memory sharing determine how fast compromise travels and how hard it is to contain. Work on malicious-agent resilience \cite{huang2024resilience}, NetSafe \cite{yu2024netsafe}, G-Safeguard \cite{wang2025g}, and communication-aware multi-agent risks \cite{hammond2025multi} supports this view. Backdoor and memory attacks make this issue more concrete. BadAgent and \emph{Watch Out for Your Agents!} show that malicious behavior can be implanted into agent workflows and activated later through interaction patterns \cite{wang2024badagent,yang2024watch}. Trojan Hippo then shows that shared memory is itself a high-risk surface, because poisoned memory can be weaponized for later exfiltration or control \cite{trojanhippo2026}. The threat moves from bad messages, to bad agents, to bad shared state. Once memory and topology are involved, rollback becomes much harder than in a single-agent system.

\subsection{Defense Agents, Attribution, and Memory Partitioning}

Recent defenses increasingly treat coordination as a security problem rather than only an efficiency problem. GuardAgent \cite{xiang2024guardagent}, AutoDefense \cite{zeng2024autodefense}, ShieldAgent \cite{chen2025shieldagent}, and related multi-agent defense pipelines \cite{multiagentdefense2025} introduce dedicated safety roles that inspect, verify, or challenge other agents before outputs are accepted. The main idea is simple: not all agents should have equal authority, and safety should be assigned explicit responsibility inside the workflow. Another important trend is attribution. Recent work asks not just whether a multi-agent system failed, but which agent, which message, and which step caused the decisive failure \cite{zhang2025agent}. This matters because containment is difficult without diagnosis. Structural defenses then push one step further. AgentSafe \cite{mao2025agentsafe} uses hierarchical data management to reduce unsafe cross-agent sharing, while G-Safeguard \cite{wang2025g} monitors the interaction graph for anomalous propagation patterns. This line of work suggests that memory partitioning, privilege separation, and topology-aware monitoring may be more durable than generic prompt filtering alone.

Overall, this boundary shows that collaboration is not automatically a safeguard. Without isolation, it can become a mechanism for amplification.

\section{System-Environment Boundary}
\label{sec:system-environment}
\subsection{Threat Model and Boundary Definition}

The system-environment boundary concerns how an agent reads and reacts to the outside world. In a well-isolated system, webpages, retrieved passages, documents, emails, interface elements, and memory artifacts should remain observations rather than hidden commands. Failure begins when environment-originated content is absorbed into the agent context and then treated as if it had authority.

The environment is often broad and weakly controlled. Once outside content can steer reasoning, retrieval, or action, the outside world becomes part of the control loop, as shown by indirect injection and retrieval-side studies \cite{greshake2023not,zverev2024can,overcomingretrieval2026,brittleagent2026}.

\subsection{Indirect Prompt Injection from Ambient Content}

The starting point of this literature is indirect prompt injection. Early work showed that malicious instructions can be hidden in webpages, emails, or documents and later executed by an LLM-integrated system when the model reads them \cite{greshake2023not}. Recent work shows that this problem is broader and more persistent than first expected. Automatic attacks \cite{liu2024automatic}, multimodal injections \cite{bagdasaryan2023abusing,wu2024wipi}, and environment-facing attacks on web agents \cite{liao2024eia,xu2024advagent} all show that the environment can steer the agent without going through the nominal user channel. \emph{Overcoming the Retrieval Barrier} \cite{overcomingretrieval2026} and \emph{Your Agent is More Brittle Than You Think} \cite{brittleagent2026} further show that indirect injection can survive realistic retrieval pipelines and remain effective even when the injected content appears peripheral. The main challenge is therefore not only detecting bad strings, but preserving source separation after retrieval and context assembly.

\subsection{Web-Agent Security and Hostile Interface Environments}

This boundary becomes even sharper in web and computer-use agents. Here the environment is not just read. It is also clicked, navigated, copied, and executed against. INJECAGENT \cite{zhan2024injecagent}, WASP \cite{wasp2025}, and WebInject \cite{webinject2025} show that browser agents are highly vulnerable when malicious prompts are embedded in realistic websites and task flows. Work on WIPI \cite{wu2024wipi}, environmental injection for privacy leakage \cite{liao2024eia}, and broader analyses of web-agent fragility \cite{chiang2025web} reaches the same conclusion from slightly different angles: web environments are active adversarial surfaces, not passive information sources. VPI-Bench \cite{vpibench2025} extends this picture to visual prompt injection, showing that computer-use agents can also be misled through interface appearance rather than text alone.

\subsection{RAG Poisoning, Retrieval Corruption, and Disclosure Risks}

RAG systems expose another major form of system-environment failure. Here the issue is often not explicit instruction, but corrupted evidence. PoisonedRAG \cite{zou2024poisonedrag}, Pandora \cite{deng2024pandora}, BADRAG \cite{xue2024badrag}, and human-imperceptible retrieval poisoning \cite{zhang2024human} show that attackers can manipulate knowledge bases or retrieved passages so that the agent reasons from compromised support. Recent work also shows that retrieval systems create disclosure risks in addition to integrity risks. Data extraction attacks \cite{peng2024data}, backdoored retrieval databases \cite{qi2024follow}, agent-based exfiltration attacks \cite{jiang2024rag}, membership inference studies \cite{li2024generating,anderson2024my}, and multimodal leakage evaluations \cite{domultimodalrag2026} show that the environment can both push corrupted knowledge in and pull private knowledge out. AgentPoison and MEMSAD further show that once retrieval outputs or memory stores are corrupted, the effect may continue across later interactions unless the system actively diagnoses and repairs the damage \cite{chen2024agentpoison,memsad2026}.

\subsection{Authentication, Provenance, and Causal Defenses}

Recent defenses increasingly try to restore explicit trust structure at this boundary. FATH \cite{wang2024fath}, Spotlighting \cite{hines2024defending}, instruction-detection methods \cite{wen2025defending,chen2025can}, and Task Shield \cite{taskshield2024} all aim to mark, isolate, or filter environment-originated instructions before they silently become control input. More recent work moves toward stronger causal and provenance-aware defenses. AgentSentry \cite{agentsentry2026} uses temporal causal diagnostics, AttriGuard \cite{attriguard2026} uses causal attribution, and TrustRAG \cite{zhou2025trustrag} emphasizes trustworthy evidence selection and robustness-aware retrieval. The broader lesson is that system-environment safety will likely depend less on stronger generic refusal and more on better source authentication, provenance tracking, memory hygiene, and context attribution.

In short, this boundary suggests that safe agent systems need not only aligned models, but also environment interfaces that preserve isolation by construction.

\section{Cross-Boundary Challenges}
\subsection{How Failures Propagate Across Boundaries}

Serious failures often cross boundaries. A common pattern is sequential escalation: user input may first override control at the user-agent boundary, then steer tool use, and finally trigger unsafe execution. Another pattern starts from the environment. A malicious webpage \cite{greshake2023not}, retrieved passage, or poisoned memory item \cite{chen2024agentpoison,trojanhippo2026} may enter through the system-environment boundary and later propagate into tool calls, agent communication \cite{lee2024prompt,he2025red}, or action traces. In multi-agent systems, the spread can continue because compromised results may be forwarded or stored for reuse \cite{he2025red,wang2025g,mao2025agentsafe}. The main unit of analysis is therefore often the full control path, not a single prompt, tool call, or action. This is also why local robustness at one interface does not guarantee system-level safety.

\subsection{Isolation-by-Construction}

Safer agent systems need interfaces that preserve separation by design. Inputs from users, tools, peer agents, and external content should remain distinguishable. Capability access should be scoped. Propagation paths should be observable through trace-level monitoring \cite{toolsafe2026}, attribution \cite{chen2025shieldagent,attriguard2026}, and policy checks \cite{agentsentry2026}. Recovery should also be a core requirement once compromise reaches memory or shared state \cite{mao2025agentsafe,memsad2026}. In practice, this means that systems should make trust, authority, and capability boundaries explicit throughout the workflow. In short, isolation-by-construction means building interfaces where the boundary remains visible to the system itself.

\begin{table*}[t]
\centering
\scriptsize
\setlength{\tabcolsep}{3.5pt}
\renewcommand{\arraystretch}{1.08}
\begin{tabular}{L{0.17\textwidth}L{0.18\textwidth}L{0.24\textwidth}L{0.34\textwidth}}
\toprule
\textbf{Prior work} & \textbf{Organizing lens} & \textbf{Main scope} & \textbf{Difference from this survey} \\
\midrule
\citet{meng2026harness} & Harness components & Agent infrastructure and engineering patterns & We reinterpret harness interfaces as safety boundaries and classify literature by the first violated contract. \\
\citet{securityconsiderations2026} & Deployment risks & System-level security considerations & We add a paper-level taxonomy separating primary isolation loss from downstream propagation. \\
\citet{landscape2026} & Attack/defense categories & Broad agentic-AI attack and defense landscape & We organize superficially different failures by their shared interface-level cause. \\
\citet{formalizing2026} & Formal security framework & Security abstractions for LLM agents & We provide broad literature mapping with a lightweight operational rule for boundary assignment. \\
\bottomrule
\end{tabular}
\caption{Positioning against representative surveys and frameworks. Our goal
is complementary: identifying where trust, authority, capability, or state
isolation first fails and how that failure propagates.}
\label{tab:survey-comparison}
\vspace{-0.15in}
\end{table*}

\subsection{Survey Protocol and Positioning}
\label{sec:survey-protocol}

\paragraph{Search and selection.}
We combined targeted searches over arXiv, ACL Anthology, DBLP, and Google
Scholar with backward and forward citation tracing. Queries paired
``LLM agent'' or ``agentic AI'' with terms covering prompt injection and
jailbreaking, tool and MCP security, code/browser/GUI execution, embodied
agents, multi-agent attacks and defenses, retrieval and memory poisoning, and
provenance. We considered work available through June 2026. We included
peer-reviewed papers and preprints that introduced or evaluated an attack,
defense, benchmark, or system design relevant to at least one agent-system
interface; duplicates, non-technical commentary, and work without a clear
agent-workflow implication were excluded. For each paper, we recorded its
first violated boundary and any downstream boundaries implicated by the
reported attack or defense. This procedure is intended to make our coverage
auditable rather than to claim exhaustiveness in a rapidly changing field.

\subsection{Operationalizing Isolation}
\label{sec:isolation-definition}

Let an agent system contain components connected by directed interfaces, each with a contract
$\Gamma_i=(T_i,A_i,C_i,S_i)$ defining admissible trust, authority, capability, and state scopes. A transfer $x$ is an \emph{isolation violation} only when its effective attributes exceed $\Gamma_i$; crossing an interface alone is not unsafe. For a causal trace $\tau=(x_1,\ldots,x_n)$, the \emph{primary boundary} is the earliest violated interface, while later affected boundaries are propagation. Violations include user content gaining control authority, tool output exceeding its declared scope, unmediated external side effects, peer contributions exceeding their roles, or external observations losing provenance and becoming trusted instructions or durable state. Thus, a poisoned webpage begins at system--environment, malicious API output at agent--tool, and generated code accessing credentials at agent--execution.

\subsection{Boundary-Aware Research Agenda}
\label{sec:boundary-agenda}

The literature is uneven: user--agent and system--environment attacks have
relatively mature benchmarks, whereas execution mediation, agent--agent
isolation, persistent-state recovery, and cross-boundary propagation remain
less systematic. Promising directions include long-session and multimodal
authority separation; typed, capability-scoped tool interfaces and authenticated
protocol metadata; reversible execution with policy checks before high-impact
actions; provenance-aware inter-agent communication, memory partitioning, and
failure attribution; and retrieval provenance, context quarantine, and memory
hygiene. Cascading-failure benchmarks should identify the initial violated
boundary, record secondary boundaries, and measure both local compromise and
eventual side effects, alongside task utility and overhead. They should also
test whether provenance survives transformations such as summarization and
retrieval, and whether a system can recover after shared or persistent state is
compromised.

\subsection{Open Challenges}

Open problems remain in evaluation, compositional defense, persistent state, and formalization. Many benchmarks still test single boundaries, while real failures are cross-boundary \cite{lee2025safearena,wasp2025,hwebench2026}. Defenses may fail after information is transformed downstream. Memory \cite{memsad2026}, retrieval caches \cite{chen2024agentpoison}, and shared agent state \cite{trojanhippo2026} further obscure and persist compromises. The field still lacks stable abstractions for authority, trust, and privilege in full workflows \cite{formalizing2026,securityconsiderations2026,landscape2026}.

\section{Conclusion}

In this survey, we argue that LLM-agent safety is best understood through boundaries. Across user input, tool use, execution, inter-agent communication, and environment interaction, failures often share a common pattern: loss of isolation at interfaces. This helps explain where compromise begins and how it propagates. Overall, safer agent systems require not only better alignment, but also clearer trust boundaries, scoped capabilities, runtime mediation, and stronger control of memory and context.

\section*{Acknowledgments}

The authors of this paper were supported by the National Key Research and Development Program of China (2025YFE0200500), the Beijing-Tianjin-Hebei Collaborative Innovation Special Project (26240201D), the ITSP Platform Research Project (ITS/189/23FP) from ITC of Hong Kong, SAR, China, and the AoE (AoE/E-601/24-N), the CRF (No. C6004-25G),  the RIF (R6021-20) and the GRF (16205322) from RGC of Hong Kong, SAR, China. The work described in this paper was conducted in full or in part by Dr. Haoran Li, JC STEM Early Career Research Fellow, supported by The Hong Kong Jockey Club Charities Trust.

\clearpage
\section*{Limitations}

This survey has several limitations. First, the field is moving quickly, so any taxonomy can become incomplete as new systems and attacks appear. Second, our taxonomy is boundary-centric by design. It is useful for isolation failures, but it is not the only valid way to organize the literature. Third, many papers are naturally cross-boundary. We classify them by the boundary where the decisive loss of isolation first occurs, but this still requires judgment. Finally, the literature is uneven across boundaries, so some parts of the survey are denser than others.

\section*{Ethical Considerations}

All authors of this paper affirm their adherence to the ACM Code of Ethics and the ACL Code of Conduct. This survey is intended to support safer research and system design for LLM agents.

At the same time, organizing the attack and defense landscape of agent systems has a dual-use aspect. We therefore focus on high-level mechanisms and design implications rather than operational attack instructions.


\clearpage
\bibliography{custom}

\clearpage
\appendix

\onecolumn
\section*{Comprehensive Summary of Representative Papers}
\input{texs/summary_table}

\end{document}

%% file: texs/taxonomy.tex
\begin{figure*}[!t]
\centering
\resizebox{\textwidth}{!}{
\begin{tikzpicture}[
    font=\tiny,
    conn/.style={draw=black!65,line width=0.5pt},
    group/.style={
        rounded corners=2pt,
        line width=0.75pt,
        minimum width=0.4cm,
        minimum height=0.1cm,
        align=center,
        font=\fontsize{6pt}{7pt}\selectfont
    },
    group2/.style={
        rounded corners=2pt,
        line width=0.75pt,
        minimum width=0.4cm,
        minimum height=0.1cm,
        align=center,
        font=\fontsize{5pt}{6pt}\selectfont
    },
    topic/.style={
        rounded corners=2pt,
        line width=0.6pt,
        minimum width=0.2cm,
        minimum height=0.1cm,
        align=center,
        inner ysep=2pt,
        font=\fontsize{5pt}{6pt}\selectfont
    },
    detail/.style={
        rounded corners=2pt,
        line width=0.6pt,
        minimum width=0.2cm,
        minimum height=0.1cm,
        align=left,
        text width=6cm,
        inner xsep=5pt,
        inner ysep=1pt,
        font=\fontsize{4pt}{5pt}\selectfont
    }
]

\def\xtrunk{0}
\def\xgroup{0.85}
\def\xbranch{1.7}
\def\xbranchh{2}
\def\xtopic{2.7}
\def\xtopicc{3.2}
\def\xdetail{7}

\def\dy{0.6}

\draw[conn] (0,-1.0) -- (0,-11.9);
\node[rotate=90,font=\bfseries\scriptsize] at (-0.3,-5.55) {Boundary-Centric Taxonomy of LLM-Agent Safety};

\def\yone{-1.0}
\node[group,draw=blue!70,fill=blue!8] (g1) at (\xgroup,\yone) {User-Agent\\Boundary\S~\ref{sec:user-agent}};
    \node[topic,draw=blue!70] (t11) at (\xtopic,{\yone+2*\dy}) {Direct prompt\\injection / jailbreaks};
    \node[topic,draw=blue!70] (t12) at (\xtopic,{\yone+\dy}) {Authority\\separation failure};
    \node[topic,draw=blue!70] (t13) at (\xtopic,\yone) {Multi-turn\\attacks};
    \node[topic,draw=blue!70] (t14) at (\xtopic,{\yone-\dy}) {Multimodal /\\long-context attacks};
    \node[topic,draw=blue!70] (t15) at (\xtopic,{\yone-2*\dy}) {Persistent\\compromise};
    \node[detail,draw=blue!70] (d11) at (\xdetail,{\yone+2*\dy}) {Direct prompt injection remains the canonical failure mode, including instruction override \cite{perez2022ignore,liu2023prompt}, jailbreak suffixes \cite{zhu2023autodan}, and adaptive attacks \cite{mazeika2024harmbench}.};
    \node[detail,draw=blue!70] (d12) at (\xdetail,{\yone+\dy}) {User content crosses privilege boundaries, behaving as policy or developer instructions rather than data.};
    \node[detail,draw=blue!70] (d13) at (\xdetail,\yone) {Compromise can emerge gradually through dialogue \cite{russinovich2024great}, contextual derailment \cite{ren2024derail}, and implicit steering \cite{zhou2024speak}.};
    \node[detail,draw=blue!70] (d14) at (\xdetail,{\yone-\dy}) {Images \cite{gong2023figstep,wu2023jailbreaking} and long contexts \cite{cheng2024leveraging} expand the attack surface beyond plain text.};
    \node[detail,draw=blue!70] (d15) at (\xdetail,{\yone-2*\dy}) {Attacks can persist through memory injection \cite{he2024data}, in-context corruption \cite{wang2024unlocking}, and residual influence \cite{xu2025mem}.};
    \draw[conn] (\xtrunk,\yone) -- (g1.west);
    \draw[conn] (g1.east) -- (\xbranch,\yone);
    \draw[conn] (\xbranch,{\yone+2*\dy}) -- (\xbranch,{\yone-2*\dy});
    \draw[conn] (\xbranch,{\yone+2*\dy}) -- (t11.west);
    \draw[conn] (\xbranch,{\yone+\dy}) -- (t12.west);
    \draw[conn] (\xbranch,\yone) -- (t13.west);
    \draw[conn] (\xbranch,{\yone-\dy}) -- (t14.west);
    \draw[conn] (\xbranch,{\yone-2*\dy}) -- (t15.west);
    \draw[conn] (t11.east) -- (d11.west);
    \draw[conn] (t12.east) -- (d12.west);
    \draw[conn] (t13.east) -- (d13.west);
    \draw[conn] (t14.east) -- (d14.west);
    \draw[conn] (t15.east) -- (d15.west);

\def\ytwo{-4}
\node[group,draw=green!60!black,fill=green!8] (g2) at (\xgroup,\ytwo) {Agent-Tool\\Boundary\S~\ref{sec:agent-tool}};
\node[topic,draw=green!60!black] (t21) at (\xtopic,{\ytwo+2*\dy}) {Tool outputs as\\control};
    \node[topic,draw=green!60!black] (t22) at (\xtopic,{\ytwo+\dy}) {Tool misuse /\\selection failures};
    \node[topic,draw=green!60!black] (t23) at (\xtopic,\ytwo) {Argument\\construction};
    \node[topic,draw=green!60!black] (t24) at (\xtopic,{\ytwo-\dy}) {Protocol /\\metadata risks};
    \node[topic,draw=green!60!black] (t25) at (\xtopic,{\ytwo-2*\dy}) {Trajectory-level\\orchestration};
    \node[detail,draw=green!60!black] (d21) at (\xdetail,{\ytwo+2*\dy}) {Tool-returned observations may be treated as trusted control rather than bounded evidence. This observation-control blur is central in indirect tool injection \cite{yi2023benchmarking} and tool feedback misuse \cite{fu2024imprompter}.};
    \node[detail,draw=green!60!black] (d22) at (\xdetail,{\ytwo+\dy}) {The system may select the wrong capability \cite{ye2024toolsword}, route toward the wrong external service \cite{zhao2024attacks}, or trust a malicious tool advertisement \cite{wang2024allies}.};
    \node[detail,draw=green!60!black] (d23) at (\xdetail,\ytwo) {Even when the correct tool is chosen, unsafe argument construction can convert local reasoning mistakes into external side effects.};
    \node[detail,draw=green!60!black] (d24) at (\xdetail,{\ytwo-\dy}) {MCP-style ecosystems show that metadata \cite{jing2025mcip}, capability descriptions and protocol manipulation \cite{wang2025mpma}, and benchmarked MCP attacks \cite{msb2025} are part of the interface attack surface.};
    \node[detail,draw=green!60!black] (d25) at (\xdetail,{\ytwo-2*\dy}) {Many failures only become visible at the trajectory level. Unsafe workflows emerge in adaptive tool traces \cite{adaptools2026}, trace-level analysis \cite{tracesafe2026}, safer invocation studies \cite{toolsafe2026}, and coding-agent benchmarks \cite{mosaic2026}.};
    \draw[conn] (\xtrunk,\ytwo) -- (g2.west);
    \draw[conn] (g2.east) -- (\xbranch,\ytwo);
    \draw[conn] (\xbranch,{\ytwo+2*\dy}) -- (\xbranch,{\ytwo-2*\dy});
    \draw[conn] (\xbranch,{\ytwo+2*\dy}) -- (t21.west);
    \draw[conn] (\xbranch,{\ytwo+\dy}) -- (t22.west);
    \draw[conn] (\xbranch,\ytwo) -- (t23.west);
    \draw[conn] (\xbranch,{\ytwo-\dy}) -- (t24.west);
    \draw[conn] (\xbranch,{\ytwo-2*\dy}) -- (t25.west);
    \draw[conn] (t21.east) -- (d21.west);
    \draw[conn] (t22.east) -- (d22.west);
    \draw[conn] (t23.east) -- (d23.west);
    \draw[conn] (t24.east) -- (d24.west);
    \draw[conn] (t25.east) -- (d25.west);

\def\ythree{-6.7}
\node[group2,draw=orange!85!black,fill=orange!10] (g3) at (\xgroup,\ythree) {Agent-Execution\\Boundary\S~\ref{sec:agent-execution}};
    \node[topic,draw=orange!85!black] (t31) at (\xtopic,{\ythree+1.5*\dy}) {Code / browser /\\GUI action risks};
    \node[topic,draw=orange!85!black] (t32) at (\xtopic,{\ythree+0.5*\dy}) {Unsafe action\\realization};
    \node[topic,draw=orange!85!black] (t33) at (\xtopic,{\ythree-0.5*\dy}) {Embodied agents /\\VLA systems};
    \node[topic,draw=orange!85!black] (t34) at (\xtopic,{\ythree-1.5*\dy}) {Containment /\\runtime mediation};
    \node[detail,draw=orange!85!black] (d31) at (\xdetail,{\ythree+1.5*\dy}) {This boundary covers the point where plans become real actions. Examples include code execution attacks \cite{zhang2024breaking}, autonomous website hacking \cite{fang2024llma}, risky code generation \cite{guo2024redcode}, and unsafe web interaction \cite{lee2025safearena}.};
    \node[detail,draw=orange!85!black] (d32) at (\xdetail,{\ythree+0.5*\dy}) {The key shift is from unsafe text to unsafe side effects: the system may click, run, submit, or manipulate the wrong thing even when the language output looks plausible.};
    \node[detail,draw=orange!85!black] (d33) at (\xdetail,{\ythree-0.5*\dy}) {In embodied and VLA systems, perception errors, backdoors, and adversarial inputs can be turned into unsafe motion or manipulation \cite{zhang2024badrobot,robey2024jailbreaking}, including adversarial VLA vulnerabilities \cite{wang2025exploringadversarial}.};
    \node[detail,draw=orange!85!black] (d34) at (\xdetail,{\ythree-1.5*\dy}) {The defense trend therefore emphasizes action containment through sandbox ecosystems \cite{zhou2024haicosystem}, policy-executable checks \cite{lu2025poex}, and realistic execution benchmarks \cite{hwebench2026}.};
    \draw[conn] (\xtrunk,\ythree) -- (g3.west);
    \draw[conn] (g3.east) -- (\xbranch,\ythree);
    \draw[conn] (\xbranch,{\ythree+1.5*\dy}) -- (\xbranch,{\ythree-1.5*\dy});
    \draw[conn] (\xbranch,{\ythree+1.5*\dy}) -- (t31.west);
    \draw[conn] (\xbranch,{\ythree+0.5*\dy}) -- (t32.west);
    \draw[conn] (\xbranch,{\ythree-0.5*\dy}) -- (t33.west);
    \draw[conn] (\xbranch,{\ythree-1.5*\dy}) -- (t34.west);
    \draw[conn] (t31.east) -- (d31.west);
    \draw[conn] (t32.east) -- (d32.west);
    \draw[conn] (t33.east) -- (d33.west);
    \draw[conn] (t34.east) -- (d34.west);

\def\yfour{-9.15}
\node[group,draw=red!70,fill=red!7] (g4) at (\xgroup,\yfour) {Agent-Agent\\Boundary\S~\ref{sec:agent-agent}};
\node[topic,draw=red!70] (t41) at (\xtopic+0.1,{\yfour+1.5*\dy}) {Prompt infection /\\communication attacks};
    \node[topic,draw=red!70] (t42) at (\xtopic,{\yfour+0.5*\dy}) {Debate /\\message propagation};
    \node[topic,draw=red!70] (t43) at (\xtopic,{\yfour-0.5*\dy}) {Topology /\\cascade / memory};
    \node[topic,draw=red!70] (t44) at (\xtopic,{\yfour-1.5*\dy}) {Defense agents /\\attribution};
    \node[detail,draw=red!70] (d41) at (\xdetail,{\yfour+1.5*\dy}) {A compromised message can become a carrier of control and spread from one agent to another. This is explicit in prompt infection \cite{lee2024prompt} and communication attacks \cite{he2025red}.};
    \node[detail,draw=red!70] (d42) at (\xdetail,{\yfour+0.5*\dy}) {Discussion, critique, debate \cite{amayuelas2024multiagent}, knowledge flooding \cite{ju2024flooding}, and communication reuse \cite{he2025red} can amplify rather than reduce malicious influence.};
    \node[detail,draw=red!70] (d43) at (\xdetail,{\yfour-0.5*\dy}) {Topology, routing, and shared memory determine whether failures stay local or become systemic. This is visible in backdoor insertion \cite{wang2024badagent}, recursive blocking \cite{zhou2025corba}, topology-aware analysis \cite{wang2025g}, and memory weaponization \cite{trojanhippo2026}.};
    \node[detail,draw=red!70] (d44) at (\xdetail,{\yfour-1.5*\dy}) {Representative defenses rely on guard agents \cite{xiang2024guardagent}, verifiable safety roles \cite{chen2025shieldagent}, failure attribution \cite{zhang2025agent}, and hierarchical data management \cite{mao2025agentsafe}.};
    \draw[conn] (\xtrunk,\yfour) -- (g4.west);
    \draw[conn] (g4.east) -- (\xbranch,\yfour);
    \draw[conn] (\xbranch,{\yfour+1.5*\dy}) -- (\xbranch,{\yfour-1.5*\dy});
    \draw[conn] (\xbranch,{\yfour+1.5*\dy}) -- (t41.west);
    \draw[conn] (\xbranch,{\yfour+0.5*\dy}) -- (t42.west);
    \draw[conn] (\xbranch,{\yfour-0.5*\dy}) -- (t43.west);
    \draw[conn] (\xbranch,{\yfour-1.5*\dy}) -- (t44.west);
    \draw[conn] (t41.east) -- (d41.west);
    \draw[conn] (t42.east) -- (d42.west);
    \draw[conn] (t43.east) -- (d43.west);
    \draw[conn] (t44.east) -- (d44.west);

\def\yfive{-11.9}
\node[group2,draw=purple!70!black,fill=purple!8] (g5) at (\xgroup,\yfive) {System-Env\\Boundary\S~\ref{sec:system-environment}};
    \node[topic,draw=purple!70!black] (t51) at (\xtopic,{\yfive+2*\dy}) {Indirect prompt\\Ambient injection};
    \node[topic,draw=purple!70!black] (t52) at (\xtopic,{\yfive+\dy}) {Hostile web /\\interfaces};
    \node[topic,draw=purple!70!black] (t53) at (\xtopic,\yfive) {RAG poisoning /\\retrieval corruption};
    \node[topic,draw=purple!70!black] (t54) at (\xtopic,{\yfive-\dy}) {Env-mediated\\disclosure};
    \node[topic,draw=purple!70!black] (t55) at (\xtopic,{\yfive-2*\dy}) {Auth / provenance /\\causal defense};
    \node[detail,draw=purple!70!black] (d51) at (\xdetail,{\yfive+2*\dy}) 
    {External content meant to be observational becomes effective control. This is the core of indirect prompt injection \cite{greshake2023not,liu2024automatic},\\including retrieval-surviving attacks \cite{overcomingretrieval2026,brittleagent2026}.};
    \node[detail,draw=purple!70!black] (d52) at (\xdetail,{\yfive+\dy}) 
    {Webpages, interfaces, and visual surfaces act as adversarial environments. Includes WIPI-style web threats \cite{wu2024wipi}, environmental privacy leakage \cite{liao2024eia}, web-agent injection \cite{wasp2025,webinject2025}, and visual injection \cite{vpibench2025}.};
    \node[detail,draw=purple!70!black] (d53) at (\xdetail,\yfive) 
    {RAG systems are vulnerable to poisoned knowledge bases \cite{zou2024poisonedrag,deng2024pandora}, retrieval corruption \cite{xue2024badrag,zhang2024human}, and compromised evidence selection turning context into faulty support.};
    \node[detail,draw=purple!70!black] (d54) at (\xdetail,{\yfive-\dy}) 
    {The same interfaces introduce confidentiality risks, including data extraction \cite{qi2024follow,jiang2024rag}, membership inference \cite{li2024generating}, and multimodal leakage \cite{domultimodalrag2026}.};
    \node[detail,draw=purple!70!black] (d55) at (\xdetail,{\yfive-2*\dy}) 
    {Defenses emphasize authentication \cite{wang2024fath}, task-alignment shielding \cite{taskshield2024}, temporal diagnosis \cite{agentsentry2026}, causal attribution \cite{attriguard2026}, and memory hygiene \cite{memsad2026}.};
    \draw[conn] (\xtrunk,\yfive) -- (g5.west);
    \draw[conn] (g5.east) -- (\xbranch,\yfive);
    \draw[conn] (\xbranch,{\yfive+2*\dy}) -- (\xbranch,{\yfive-2*\dy});
    \draw[conn] (\xbranch,{\yfive+2*\dy}) -- (t51.west);
    \draw[conn] (\xbranch,{\yfive+\dy}) -- (t52.west);
    \draw[conn] (\xbranch,\yfive) -- (t53.west);
    \draw[conn] (\xbranch,{\yfive-\dy}) -- (t54.west);
    \draw[conn] (\xbranch,{\yfive-2*\dy}) -- (t55.west);
    \draw[conn] (t51.east) -- (d51.west);
    \draw[conn] (t52.east) -- (d52.west);
    \draw[conn] (t53.east) -- (d53.west);
    \draw[conn] (t54.east) -- (d54.west);
    \draw[conn] (t55.east) -- (d55.west);

\end{tikzpicture}
}
\caption{A roadmap-style view of our boundary-centric taxonomy for LLM-agent system safety. Each boundary is organized into representative subtopics and example literature.}
\label{fig:roadmap-taxonomy}
\vspace{-0.2in}
\end{figure*}
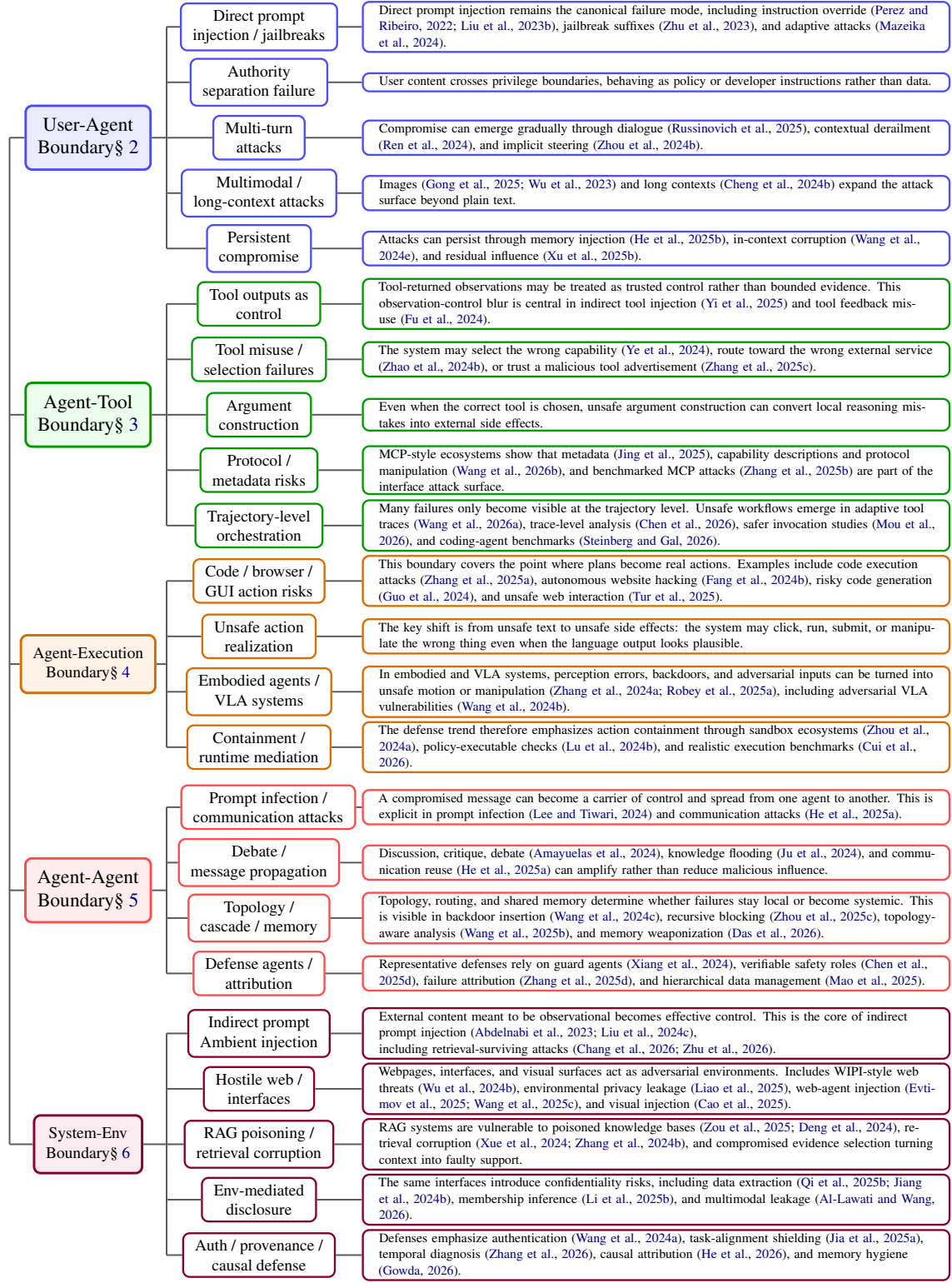

%% file: texs/summary_table.tex
\renewcommand{\arraystretch}{1.03}
\setlength{\tabcolsep}{3.5pt}

\scriptsize
\begin{longtable}{L{3.0cm} L{0.65cm} L{1.45cm} L{2.35cm} L{7.0cm}}
\caption{Comprehensive summary of representative papers in our survey. Papers are grouped by their primary boundary, even when some works naturally span multiple boundaries.}
\label{tab:paper_master_summary}\\

\toprule
\rowcolor{black!6}
\textbf{Paper} & \textbf{Year} & \textbf{Type} & \textbf{Subtopic} & \textbf{Main Idea} \\
\endfirsthead

\toprule
\rowcolor{black!6}
\textbf{Paper} & \textbf{Year} & \textbf{Type} & \textbf{Subtopic} & \textbf{Main Idea} \\
\midrule
\endhead

\midrule
\multicolumn{5}{r}{\footnotesize Continued on next page} \\
\endfoot

\bottomrule
\endlastfoot

\boundarysection{blue}{User-Agent Boundary}

\cite{perez2022ignore} & 2022 & Attack & Prompt injection & Early evidence that language instructions can override hidden prompts and expose role confusion. \\
\cite{liu2023prompt} & 2023 & Attack & Prompt injection & Shows that system prompts can be leaked or overridden through ordinary user interaction. \\
\cite{toyer2023tensor} & 2023 & Attack & Prompt injection & Studies prompt-injection style attacks against tool-using language systems. \\
\cite{pedro2023prompt} & 2023 & Attack & Role confusion & Shows how hidden instructions and visible user content can collapse into the same control channel. \\
\cite{zhu2023autodan} & 2023 & Attack & Automated jailbreaks & Introduces optimization-based jailbreak generation at scale rather than only handcrafted prompts. \\
\cite{liu2023autodan} & 2023 & Attack & Automated jailbreaks & Develops automated jailbreak generation with stronger transfer and search efficiency. \\
\cite{huang2023catastrophic} & 2024 & Attack & Jailbreaks & Shows catastrophic jailbreak behavior in open-weight models through generation exploitation. \\
\cite{chao2023jailbreaking} & 2023 & Attack & Black-box jailbreaks & Demonstrates efficient black-box jailbreaking with limited interaction budget. \\
\cite{wei2023jailbreak} & 2023 & Attack & In-context jailbreaks & Shows that few-shot demonstrations can jailbreak or guard aligned language models. \\
\cite{jia2024improved} & 2024 & Attack & Optimization attacks & Improves optimization-based jailbreak methods against aligned models. \\
\cite{sitawarin2024pal} & 2024 & Attack & Black-box attacks & Uses proxy-guided attack strategies to strengthen jailbreak transfer. \\
\cite{jiang2024dartdeepadversarialautomated} & 2024 & Attack & Automated red teaming & Uses deep adversarial search for automated safety testing. \\
\cite{pei2024selfprompt} & 2024 & Attack & Automated red teaming & Uses self-generated prompts to test robustness under constrained attack settings. \\
\cite{mazeika2024harmbench} & 2024 & Benchmark & Evaluation & Provides a broad benchmark for harmful behavior and jailbreak evaluation. \\
\cite{chao2024jailbreakbench} & 2024 & Benchmark & Evaluation & Standardizes jailbreak testing across models and attack styles. \\
\cite{ran2024jailbreakeval} & 2024 & Benchmark & Evaluation & Provides systematic evaluation of jailbreak effectiveness and defense performance. \\
\cite{russinovich2024great} & 2024 & Attack & Multi-turn attacks & Shows that compromise may emerge gradually through repeated dialogue rather than a single prompt. \\
\cite{ren2024derail} & 2024 & Attack & Multi-turn attacks & Studies derailment over extended interaction rather than one-shot failure. \\
\cite{cheng2024leveraging} & 2024 & Attack & Long-context attacks & Shows that long context can be used to weaken or bypass safety behavior. \\
\cite{chang2024play} & 2024 & Attack & Indirect clues & Uses implicit clues and guessing-game style interaction for indirect jailbreak. \\
\cite{upadhayay2024cognitive} & 2024 & Attack & Long-context attacks & Uses cognitive overload in long context to weaken safety behavior. \\
\cite{gong2023figstep} & 2023 & Attack & Multimodal attacks & Demonstrates that visual channels can bypass text-focused safety assumptions. \\
\cite{liu2023query} & 2023 & Attack & Multimodal attacks & Shows that query-relevant images can jailbreak multimodal systems. \\
\cite{wu2023jailbreaking} & 2023 & Attack & Multimodal attacks & Shows that multimodal inputs can be used to induce unsafe behavior. \\
\cite{he2024data} & 2024 & Attack & Persistence & Shows that in-context state can be poisoned and later reused. \\
\cite{wang2024unlocking} & 2024 & Attack & Persistence & Studies latent unlocking of unsafe behavior through hidden contextual influence. \\
\cite{xu2025mem} & 2025 & Attack & Memory injection & Shows that memory can preserve unsafe user influence across interactions. \\
\cite{chen2024struq} & 2024 & Defense & Authority separation & Uses structured prompting to separate instructions from data more explicitly. \\
\cite{suo2024signed} & 2024 & Defense & Authority separation & Uses signed prompting or explicit authority structure to preserve source distinction. \\
\cite{sharma2024spml} & 2024 & Defense & DSL-style defense & Uses a domain-specific language to separate trusted control from untrusted prompt content. \\
\cite{robey2023smoothllm} & 2023 & Defense & Jailbreak defense & Proposes smoothing-style robustness techniques against prompt attacks. \\
\cite{kim2023robust} & 2023 & Defense & Safety classifier & Uses robust safety classification as a shield against adversarial prompting. \\
\cite{ji2024defending} & 2024 & Defense & Semantic smoothing & Defends against jailbreak attacks through semantic smoothing techniques. \\
\cite{wang2024selfdefend} & 2024 & Defense & Inference-time defense & Uses self-protection strategies at inference time against unsafe prompting. \\
\cite{lin2025uniguardian} & 2025 & Defense & Inference-time defense & Uses model-side protection at inference time against unsafe user control. \\
\cite{ying2026agentvisor} & 2026 & Defense & Semantic virtualization & Uses semantic virtualization to defend agents against prompt injection through stronger separation of trusted and untrusted context. \\

\boundarysection{green}{Agent-Tool Boundary}

\cite{yi2023benchmarking} & 2023 & Attack & Tool outputs as control & Shows that indirect prompt injection can arise when tool-returned content is treated as trusted instruction. \\
\cite{ye2024toolsword} & 2024 & Analysis & Tool misuse & Breaks tool-use failure into interpretation, selection, and execution stages. \\
\cite{fu2024imprompter} & 2024 & Analysis & Tool feedback misuse & Studies how tool-learning and feedback use can create new safety failures. \\
\cite{zhao2024attacks} & 2024 & Attack & Tool selection & Shows that external tools and APIs can be manipulated before or during invocation. \\
\cite{wang2024allies} & 2024 & Attack & Adversarial tool-calling & Demonstrates that the right tool can still be used in the wrong way. \\
\cite{attacktoolselection2025} & 2025 & Attack & Tool selection & Studies prompt injection and adversarial interference against tool selection logic. \\
\cite{jing2025mcip} & 2025 & Attack & MCP / protocol risk & Shows that capability descriptions and protocol metadata are themselves safety-critical. \\
\cite{wang2025mpma} & 2025 & Attack & Metadata manipulation & Studies manipulation through protocol-level capability exposure and routing signals. \\
\cite{msb2025} & 2025 & Benchmark & MCP security & Benchmarks attacks against model context protocol ecosystems. \\
\cite{toolsafe2026} & 2026 & Defense & Safer invocation & Focuses on stronger mediation for high-risk tool calls. \\
\cite{chen2025agentguard} & 2025 & Defense & Invocation control & Uses guard mechanisms to mediate tool access and high-impact external capabilities. \\
\cite{adaptools2026} & 2026 & Attack & Trace-level risk & Studies multi-step unsafe tool orchestration rather than single calls. \\
\cite{tracesafe2026} & 2026 & Defense & Trace-level monitoring & Emphasizes safety over full tool-calling trajectories. \\
\cite{ciber2026} & 2026 & Benchmark & Coding agents & Evaluates code-interpreter style agents under tool-mediated risk. \\
\cite{mosaic2026} & 2026 & Benchmark & Coding agents & Measures compositional vulnerability in tool-heavy coding workflows. \\

\boundarysection{orange}{Agent-Execution Boundary}

\cite{zhang2024breaking} & 2024 & Attack & Unsafe execution & Shows that autonomous agents can be compromised into harmful behavior through malfunction amplification. \\
\cite{fang2024llma} & 2024 & Attack & Web exploitation & Demonstrates that LLM agents can autonomously hack websites. \\
\cite{fang2024llmb} & 2024 & Attack & Vulnerability exploitation & Extends execution risk to one-day vulnerability exploitation. \\
\cite{guo2024redcode} & 2024 & Benchmark & Code agents & Benchmarks risky code generation and execution behavior. \\
\cite{shlomov2024st} & 2024 & Benchmark & Web agents & Evaluates safety and trustworthiness in web-agent interaction traces. \\
\cite{lee2025safearena} & 2025 & Benchmark & Web agents & Evaluates safety in autonomous web-agent action sequences. \\
\cite{kumar2025refusal} & 2025 & Analysis & Browser agents & Shows that refusal-trained models remain vulnerable once coupled to browser action. \\
\cite{chen2025obvious} & 2025 & Attack & GUI agents & Highlights hidden threats in LLM-powered GUI control. \\
\cite{zhang2024badrobot} & 2024 & Attack & Embodied agents & Demonstrates jailbreaking of embodied systems in the physical world. \\
\cite{robey2024jailbreaking} & 2024 & Attack & Robot control & Studies prompt-based compromise in robot systems controlled by LLMs. \\
\cite{liu2024compromising} & 2024 & Attack & Embodied backdoors & Shows contextual backdoor attacks against embodied agents. \\
\cite{jiao2025canwetrust} & 2025 & Attack & Embodied backdoors & Studies whether embodied decision-making agents can be trusted under backdoor attack. \\
\cite{hu2024viva} & 2024 & Benchmark & Vision-grounded safety & Benchmarks decision making with human-value constraints in embodied settings. \\
\cite{li2024embodiedagentinterface} & 2024 & Benchmark & Embodied evaluation & Benchmarks embodied decision making through a dedicated agent interface. \\
\cite{tomilin2025hasard} & 2025 & Benchmark & Safe RL / embodiment & Benchmarks safe embodied decision making under visual risk. \\
\cite{chakraborty2025heal} & 2025 & Analysis & Embodied hallucination & Studies hallucinations in embodied agents driven by LLMs. \\
\cite{karnik2025embodiedredteaming} & 2025 & Benchmark & Embodied red teaming & Audits robotic foundation models through embodied red teaming. \\
\cite{wu2024adversarial} & 2024 & Attack & Multimodal agents & Studies adversarial attacks on multimodal agents more broadly. \\
\cite{zhou2025exploringlimits} & 2025 & Attack & Cross-task manipulation & Explores manipulation limits and generalization in VLA systems. \\
\cite{wang2025advancingembodied} & 2025 & Defense & Embodied moderation & Connects safety benchmarks to input moderation in embodied agents. \\
\cite{wang2025exploringadversarial} & 2025 & Attack & VLA vulnerabilities & Explores adversarial weaknesses in vision-language-action models. \\
\cite{cheng2024manipulationfacingthreats} & 2024 & Benchmark & Physical vulnerability & Evaluates physical vulnerabilities in end-to-end VLA manipulation tasks. \\
\cite{zhou2024haicosystem} & 2024 & Defense & Sandboxing & Frames execution safety as a containment and sandboxing problem. \\
\cite{lu2025poex} & 2025 & Defense & Policy-executable defense & Proposes executable safeguards against embodied jailbreaks. \\
\cite{2603.17419} & 2026 & Defense & Zero-trust runtime & Uses a zero-trust architecture to constrain autonomous execution in high-stakes settings. \\
\cite{hwebench2026} & 2026 & Benchmark & Hardware repair & Evaluates long-horizon execution safety in real repair tasks. \\
\cite{fokou2026parallax} & 2026 & Defense & Reason-act separation & Argues that systems that reason should not directly act without stronger separation and execution mediation. \\

\boundarysection{red}{Agent-Agent Boundary}

\cite{lee2024prompt} & 2024 & Attack & Prompt infection & Shows that one agent can pass malicious control content to another. \\
\cite{amayuelas2024multiagent} & 2024 & Attack & Debate attacks & Studies adversarial influence through multi-agent debate. \\
\cite{ju2024flooding} & 2024 & Attack & Knowledge propagation & Shows that manipulated knowledge can spread socially across agent communities. \\
\cite{he2025red} & 2025 & Attack & Communication attacks & Red-teams multi-agent systems by targeting communication channels directly. \\
\cite{wang2024badagent} & 2024 & Attack & Backdoored agents & Studies insertion and activation of backdoors in agent workflows. \\
\cite{yang2024watch} & 2024 & Attack & Backdoor threats & Highlights latent malicious behavior in LLM-based agents. \\
\cite{zhou2025corba} & 2025 & Attack & Cascade failure & Studies contagious recursive blocking and communication-level collapse. \\
\cite{trojanhippo2026} & 2026 & Attack & Shared memory & Shows that agent memory can be weaponized for later exfiltration. \\
\cite{huang2024resilience} & 2024 & Analysis & Malicious agents & Studies resilience of multi-agent systems when some agents behave maliciously. \\
\cite{yu2024netsafe} & 2024 & Analysis & Topology safety & Shows that network structure shapes safety and spread in multi-agent communities. \\
\cite{hammond2025multi} & 2025 & Analysis & Systemic risk & Studies broader multi-agent risks from advanced AI systems. \\
\cite{wang2025g} & 2025 & Defense & Topology monitoring & Uses interaction graphs to study and mitigate system-level spread. \\
\cite{mao2025agentsafe} & 2025 & Defense & Memory partitioning & Uses hierarchical data management to reduce unsafe cross-agent sharing. \\
\cite{xiang2024guardagent} & 2024 & Defense & Guard roles & Assigns a dedicated guard agent to inspect other agents' behavior. \\
\cite{zeng2024autodefense} & 2024 & Defense & Defense agents & Uses a multi-agent defense setup against jailbreak attacks. \\
\cite{chen2025shieldagent} & 2025 & Defense & Verifiable policy reasoning & Adds explicit reasoning about safety policy within agent coordination. \\
\cite{zhang2025agent} & 2025 & Analysis & Failure attribution & Identifies which agent and which step caused a task failure. \\
\cite{multiagentdefense2025} & 2025 & Defense & Prompt-injection defense & Uses a dedicated multi-agent defense pipeline against prompt injection. \\

\boundarysection{purple}{System-Environment Boundary}

\cite{greshake2023not} & 2023 & Attack & Indirect prompt injection & Establishes that external content can later act as hidden control in LLM systems. \\
\cite{liu2024automatic} & 2024 & Attack & Automatic injection & Scales indirect prompt injection through more systematic attack generation. \\
\cite{bagdasaryan2023abusing} & 2023 & Attack & Multimodal injection & Shows that images and sounds can act as indirect instruction channels. \\
\cite{wu2024wipi} & 2024 & Attack & Web threats & Shows new web-specific threats for LLM-driven web agents. \\
\cite{liao2024eia} & 2024 & Attack & Privacy leakage & Studies environmental injection that induces privacy leakage in web agents. \\
\cite{xu2024advagent} & 2024 & Attack & Web-agent red teaming & Uses controllable black-box red teaming against web agents. \\
\cite{zverev2024can} & 2024 & Analysis & Instruction-data separation & Examines whether LLMs can reliably separate instructions from data. \\
\cite{overcomingretrieval2026} & 2026 & Attack & Retrieval-mediated injection & Shows that indirect injection can survive realistic retrieval pipelines. \\
\cite{brittleagent2026} & 2026 & Attack & Long-horizon fragility & Shows that agentic systems remain vulnerable even when hostile content appears peripheral. \\
\cite{zhan2024injecagent} & 2024 & Benchmark & Web agents & Benchmarks indirect prompt injection in tool-integrated agents. \\
\cite{wasp2025} & 2025 & Benchmark & Web agents & Benchmarks prompt-injection robustness of web agents. \\
\cite{webinject2025} & 2025 & Attack & Web prompt injection & Specializes prompt injection to web-agent interaction. \\
\cite{vpibench2025} & 2025 & Benchmark & Visual injection & Studies visual prompt injection against computer-use agents. \\
\cite{zou2024poisonedrag} & 2024 & Attack & RAG poisoning & Studies knowledge corruption attacks against retrieval-augmented generation. \\
\cite{deng2024pandora} & 2024 & Attack & RAG jailbreaks & Shows that retrieval poisoning can induce jailbreak-like behavior. \\
\cite{xue2024badrag} & 2024 & Attack & Retrieval corruption & Identifies vulnerabilities in RAG systems under corrupted retrieval. \\
\cite{zhang2024human} & 2024 & Attack & Retrieval poisoning & Shows that imperceptible retrieval poisoning can alter downstream behavior. \\
\cite{peng2024data} & 2024 & Attack & Backdoored extraction & Uses backdoors to extract data from retrieval-augmented systems. \\
\cite{qi2024follow} & 2024 & Attack & Data extraction & Demonstrates scalable data extraction from RAG systems. \\
\cite{jiang2024rag} & 2024 & Attack & Agent-based exfiltration & Uses agent workflows to extract private data from RAG systems. \\
\cite{li2024generating} & 2024 & Attack & Membership inference & Studies whether sensitive data can be inferred from RAG outputs. \\
\cite{anderson2024my} & 2024 & Attack & Membership inference & Asks whether specific data exists in the retrieval database. \\
\cite{chen2024agentpoison} & 2024 & Attack & Memory poisoning & Shows that agent memory or knowledge bases can be poisoned for later exploitation. \\
\cite{domultimodalrag2026} & 2026 & Attack & Multimodal leakage & Extends retrieval leakage analysis to multimodal RAG settings. \\
\cite{wang2024fath} & 2024 & Defense & Authentication & Uses authentication-style defenses against indirect prompt injection. \\
\cite{hines2024defending} & 2024 & Defense & Spotlighting & Makes suspicious external instructions more visible during inference. \\
\cite{wen2025defending} & 2025 & Defense & Instruction detection & Uses instruction detection to filter indirect prompt injection. \\
\cite{taskshield2024} & 2024 & Defense & Task alignment & Enforces task alignment against hostile external instructions. \\
\cite{chen2025can} & 2025 & Defense & Detection / removal & Studies whether indirect prompt injection can be detected and removed after ingestion. \\
\cite{agentsentry2026} & 2026 & Defense & Temporal diagnosis & Tracks when environment content begins to dominate later behavior. \\
\cite{attriguard2026} & 2026 & Defense & Causal attribution & Uses attribution to separate benign context from attack-driving context. \\
\cite{memsad2026} & 2026 & Defense & Memory hygiene & Detects memory poisoning in retrieval-augmented agents. \\
\cite{formalizing2026} & 2026 & Analysis & Formalization & Provides a framework for formalizing LLM agent security beyond isolated attack cases. \\
\cite{securityconsiderations2026} & 2026 & Analysis & Systems security & Summarizes system-level security considerations for AI agents deployed in real environments. \\
\cite{landscape2026} & 2026 & Survey & Threat landscape & Synthesizes the broader 2026 attack and defense landscape of agentic AI. \\
\cite{zhou2025trustrag} & 2025 & Defense & Trustworthy retrieval & Emphasizes trustworthy evidence selection and robustness-aware design in RAG systems. \\

\end{longtable}
\normalsize